# A CNN for homogneous Riemannian manifolds with applications to Neuroimaging

Rudrasis Chakraborty[†], Monami Banerjee[†] and Baba C. Vemuri [*]
Department of CISE, University of Florida, FL 32611, USA
[†] Equal Contribution

{rudrasischa, monamie.b, baba.vemuri}@gmail.com

August 6, 2018

## Abstract

Convolutional neural networks are ubiquitous in Machine Learning applications for solving a variety of problems. They however can not be used in their native form when the domain of the data is commonly encountered manifolds such as the sphere, the special orthogonal group, the Grassmanian, the manifold of symmetric positive definite matrices and others. Most recently, generalization of CNNs to data domains such as the 2-sphere has been reported by some research groups, which is referred to as the spherical CNNs (SCNNs). The key property of SCNNs distinct from CNNs is that they exhibit the rotational equivariance property that allows for sharing learned weights within a layer. In this paper, we theoretically generalize the CNNs to Riemannian homogeneous manifolds, that include but are not limited to the aforementioned example manifolds. Our key contributions in this work are: (i) A theorem stating that linear group equivariance systems are fully characterized by correlation of functions on the domain manifold and vice-versa. This is fundamental to the characterization of all linear group equivariant systems and parallels the widely used result in linear system theory for vector spaces. (ii) As a corrolary, we prove the equivariance of the correlation operation to group actions admitted by the input domains which are Riemannian homogeneous manifolds. (iii) We present the first end-to-end deep network architecture for classification of diffusion magnetic resonance image (dMRI) scans acquired from a cohort of 44 Parkinson Disease patients and 50 control/normal subjects. (iv) A proof of concept experiment involving synthetic data generated on the manifold of symmetric positive definite matrices is presented to demonstrate the applicability of our network to other types of domains.

[*]This research was funded in part by the NSF grant IIS-1525431 and IIS-1724174 to BCV.

# 1 Introduction

CNNs introduced by Lecun [19] have gained enormous attention in the past decade especially after the demonstration of the significant success on Imagenet data by Krizhevsky et al. [18] and others. The key property of equivariance to translation of patterns in the image is utilized in the CNN to share learned weights across a layer in the network. Thus, one might consider exploiting equivariance to transformation groups as a key design principle in designing neural network architectures suitable for these groups. For data sets that are samples of functions defined on Riemannian manifolds, it would then be natural to seek a symmetry group property that the manifold admits and define the correlation operation (on the manifold) that would be equivariant to this symmetry group. For instance, on the n-sphere, the natural group action is the rotation group. Rotations in n-dimensions are elements of the well known group called the special orthogonal group, $\mathsf{SO}(n)$, which is a Lie group [14]. Thus, to develop correlation of functions on the sphere, one seeks equivariance with respect to rotations. This way, one has the flexibility to seek filters/masks that are orientation sensitive.

Steerable filters have been recognized as being of great importance in Computer Vision literature several decades ago in the context of hand crafted features [13, 9, 22]. Steerability here refers to synthesis of all deformations (scale, rotation and translation) of the filter using a small number of basis functions. Recently, steerable CNNs have been reported in literature [10, 7, 6] that are applicable to data which are samples of functions on a 2-sphere and hence are equivariant with respect to rotations in 3D. In [11], authors describe what they call polar transformer networks, which are equivariant to rotations and scaling transformations defined on the domain of the data. By combining them with what is known in literature as a spatial transformer [16], they achieve the required equivariance to translations as well. For literature on deep networks where data reside on 2-manifolds, we refer the interested reader to a recent excellent survey paper [2] and references therein.

In this paper, we present a generalization to the well known and widely used result that linear shift-equivariant systems are characterized fully by convolution and vice-versa in linear system theory for vector-spaces. In the context of deep networks, the actual operation is not a convolution but a correlation [7]. We will therefore use the term correlation instead of convolution from here on for the rest of the paper. The generalization here states that linear group equivariant systems on these manifolds are characterized by correlation and vice-versa. As a corrolary, we prove that correlation of functions has the property of equivariance to group actions admitted by Riemannian homogeneous manifolds. A homogeneous Riemannian manifold for a group $G$ is a nonempty topological space $\mathcal{M}$ on which $G$ acts transitively [15]. The elements of $G$ are called symmetries of $\mathcal{M}$. Intuitively, a homogeneous space, a.k.a. a homogeneous Riemannian manifold, is one which 'looks" the same locally at all points on it with respect to some geometric property such as isometry, diffeomorphism etc. Some important examples of homogeneous spaces include Riemannian symmetric spaces e.g., the n-sphere ($\mathbf{S}^n$), projective space, Euclidean space, the Grassmanian, manifold of $(n, n)$ symmetric positive definite matrices denoted by $P_n$, and others. Most of these manifolds are commonly encountered in mathematical formulations of various Computer Vision tasks such as action recognition, covariance tracking etc. and in Medical Imaging for example in dMRI, elastography, conductance imaging etc. By providing a general framework suited for data on non-Euclidean domains, that is analogous to the CNNs, in this paper, we extend the success of CNNs to several high dimensional yet unexplored input data domains. At the time of submission of this paper, we were made aware of recent work in [17] on a generalization of CNNs to homogeneous spaces. We acknowledge their work as being parallel work but also would like to point out significant distinctions between our work here and theirs. (i) Their main result on equivariance of correlations to group actions admitted by the homogeneous space is stated and proved for the entire deep network unlike ours that is stated as a corrolary to a more general theorem on characterization of linear group-equivariant systems and proved to hold within a layer. Note that in any CNN, equivariance to group action should be exhibited within a layer and not across the layers since there are nonlinear units between layers that will invalidate the equivariance property. (ii) They assume the total space in the definition of the homogeneous space to be compact, a restriction we do not make. Our proof of equivariance is very distinct from theirs and simpler. (iii) We present a novel architecture for implementing the HCNN and include experiments depicting for the first time an end-to-end implementation for classification of dMRI brain scans of Parkinson Disease patients and control subjects.

We present two sets of experiments to demonstrate the performance of our theoretical framework that will henceforth be called, the HCNN for homogeneous space CNN. The first experiment involves synthetic data where we synthesize two classes consisting of functions on $P_3$. We classify data samples drawn from these two classes of distributions using our proposed HCNN framework.

Our second experiment is on real data, involving classification of dMRI scans acquired from a cohort of 44 Parkinson Disease (PD) patients and 50 Control subjects. We present two approaches to solve this classification problem. In the first, to the best of our knowledge, we present the first end-to-end classification of dMRI brain scans. dMRI is a non-invasive magnetic resonance imaging technique that allows for inference of neuronal connectivity between various neuroanatomical structures using a diffusion sensitized MR signal [1]. Typically, diffusion sensitizing magnetic field gradients are applied along a large number (typically 64) of directions spanning the hemisphere of directions and the response MR signal is collected at each voxel along these directions. For each direction, the data contains an entire MR volume image. Let $S(\mathbf{q})$, denote the scalar valued signal at a voxel in the 3D image along a radial vector $\mathbf{q}$ in the Fourier (frequency) space. Note that MR image acquisition is performed in the Fourier space. Since $\mathbf{q}$ is a radial vector, the natural mathematical space for representing this signal $S(\mathbf{q})$ is then by functions on the product space, $\mathbf{S}^2 \times \mathbf{R}^+$. This product space is a Riemannian homogeneous space. Using the spherical harmonic basis on $\mathbf{S}^2$ – which form a basis for all $L^2$ functions on the sphere – along with Laguerre polynomials for representing the radial part, we present group equivariant correlation for such functions. The product is known by several names in literature, the Fourier-Laguerre basis [20], the SHORE basis in dMRI literature [21, 12]. In the following sections, we will present the theory and implementation along with experiments for this setting.

The second approach to this problem involves, first computing the ensemble average propagator (EAP), a probability density function $P(\mathbf{r})$ from the raw dMRI data. The EAP quantifies the probability of a water molecule moving along the radial vector $\mathbf{r} \in \mathbf{R}^3$. The EAP is known to fully capture the intr-voxel diffusional characteristics in the image lattice. The EAP is related to the raw diffusion sensitized magnetic resonance (MR) signal, $S(\mathbf{q})$, via the 3D Fourier transform $P(\mathbf{r}) = \int S(\mathbf{q}) \exp 2\pi j \mathbf{q}^t \mathbf{r} d\mathbf{r}$ [3]. We represent each EAP as a function on the product manifold, $\mathbf{R}^+ \times S^2$. This provides us with data similar to the one in the first approach except in the Fourier dual space. Subsequently, we present experiments on classification in this setting to compare with the first approach described above. In all of our experiments, we have shown high classification accuracy using our proposed HCNN framework.

In summary, our key contributions in this paper are: (i) We prove that any linear group equivariant system is characterized by a correlation and vice-versa. (ii) As a corollary, we prove the equivariant property of correlation to group actions admitted by the homogeneous manifold. (iii) We present a novel deep architecture for implementing HCNN. (iv) To the best of our knowledge, for the first time, an end-to-end implementation for classification of dMRI brain scans acquired from PD patients and controls is presented. We also present a synthetic data example for classification of data on $P_3$.

Rest of the paper is organized as follows. In section 2 we present the main theoretical results on equivariance of convolution on homogeneous spaces and the linear-group-equivariant systems. Section 3 contains a detailed description of the proposed HCNN architecture. In section 4 we present the experimental results and conclude in section 5.

## 2 Correlation on Riemannian homogeneous spaces

In this section, we first briefly give an overview on the differential geometry of Riemannian homogeneous spaces required in the rest of the paper. For a detailed exposition on these concepts, we refer the reader to a comprehensive and excellent treatise by Helgason [15]. Then, we define correlation of functions on the input domain which is a homogeneous space. Following which, we will prove the main theorem of this paper that is analogous to the widely known result in linear systems theory in vector spaces namely, *any linear group equivariant system can be expressed by a correlation operation*. A corollary of this theorem is the group equivariance property of correlation.

**Definition 1** (**Riemannian homogeneous space**). *Let $(\mathcal{M}, g^{\mathcal{M}})$ be a Riemannian manifold with a Riemannian metric $g^{\mathcal{M}}$. Let $G = I(\mathcal{M})$ be the set of isometries acting transitively on $\mathcal{M}$ and $H = Stab(o)$, $o \in \mathcal{M}$ (called the "origin" of $\mathcal{M}$) be a subgroup of $G$, where Stab is the Stabilizer. Then, $\mathcal{M}$ is a* homogeneous space and can be identified with the quotient space $G/H$ under the diffeomorphic mapping $gH \mapsto g.o, g \in G$ [15], where . is the group action.

For the rest of the paper, we will use the term homogeneous space to denote a Riemannian homogeneous space equipped with a Riemannian metric. Below, we present some of the properties of homogeneous spaces that we need in the rest of the paper.

**Properties of Homogeneous spaces:** Let $(\mathcal{M}, g^{\mathcal{M}})$ be a Homogeneous space. Let $\omega^{\mathcal{M}}$ be the corresponding volume density and $f : \mathcal{M} \to \mathbf{R}$ be any integrable function. Let $g \in G$, s.t. $y = g.x$, $x, y \in \mathcal{M}$. Then, the following facts are true:

1. $g^{\mathcal{M}}(dy, dy) = g^{\mathcal{M}}(dx, dx)$.

2. $d(x, z) = d(y, g.z)$, for all $z \in \mathcal{M}$.

3. $\int_{\mathcal{M}} f(y) \omega^{\mathcal{M}}(x) = \int_{\mathcal{M}} f(x) \omega^{\mathcal{M}}(x)$

Because of property 1, the volume density is also preserved by the transformation $x \mapsto g.x$, for all $x \in \mathcal{M}$ and $g \in G$.

Given two $L^2$ (square integrable) functions $f : \mathcal{M} \to \mathbf{R}$ and $w : \mathcal{M} \to \mathbf{R}$, the correlation is defined as:

**Definition 2** (**Correlation**). *Using the above notations, the correlation between $f$ and $w$ is given by, $(f \star w) : G \to \mathbf{R}$ as follows:*

$$(f \star w)(g) := \int_{\mathcal{M}} f(x) \left(L_*^g w\right)(x) \omega^{\mathcal{M}}(x), \tag{1}$$

where $L^g$ is the left group action and $L_*^g w$ is the pushforward of $w$ with the diffeomorphism $L^g$. Before proving the main theorem of this paper, we will first define a linear group equivariant system. In the rest of the paper we will assume $\mathcal{M}$ is a homogeneous space on which $G$ acts transitively. Furthermore, we will always assume functions are square-integrable. Equivariance to the group operation admitted by the homogeneous space is defined as follows:

**Definition 3** (**Equivariance**). *Let $X$ and $Y$ be $G$ sets [14]. Then, $F : X \to Y$ is said to be equivariant if*

$$F(g.x) = g.F(x) \tag{2}$$

*for all $g \in G$ and all $x \in X$.*

Let $S = \{f : \mathcal{M} \to \mathbf{R}\}$. Then, $S$ is a G set.

**Proposition 1.** *Let $w \in S$. Let $U = \{(f \star w) : G \to \mathbf{R} | f \in S\}$. Then, $U$ is a $G$ set.*

*Proof.* Let, $g_1 \in G$, $(g.(f \star w))(g_1) = (f \star w)(g^{-1} g_1) = \int_{\mathcal{M}} f(x) w(g_1^{-1} g.x) \omega^{\mathcal{M}}(x)$ $= \int_{\mathcal{M}} f(g^{-1}.y) w(g_1^{-1} y) \omega^{\mathcal{M}}(y) = \int_{\mathcal{M}} h(y) w(g_1^{-1} y) \omega^{\mathcal{M}}(y)$. Here, $h \in S$ such that, $h(y) = f(g^{-1}.y)$. As, $g_1$ is arbitrary, hence $U$ is a $G$ set. ∎

**Definition 4** (**Linear group equivariant (LGE) system**). *Let $S$ and $U$ be $G$ sets of square-integrable functions. A function $F : S \to U$ is a linear group equivariant function* iff

- *$F$ is linear, i.e., $F(c_1 f_1 + c_2 f_2) = c_1 F(f_1) + c_2 F(f_2)$.*

- *$F$ is equivariant as defined in Definition 3.*

We will now prove the theorem which gives the equivalence of linear equivariant system and the correlation operator defined above.

**Theorem 1.** *Let $F : S \to U$ be a linear equivariant function. Then, $\exists w \in S$ such that, $(F(f))(g) = (f \star w)(g)$, for all $f \in S$ and $g \in G$.*

*Proof.* Let $f \in S$ and $g \in G$. Let $x \in \mathcal{M}$, then, $f(x) = \int_{\mathcal{M}} f(y) u_y(x) \omega^{\mathcal{M}}(y)$, where, $u_y : \mathcal{M} \to \mathbf{R}$ is zero everywhere except at $x \in \mathcal{M}$ when $d(x,y) = 0$. $u_y$ is analogous to the delta function. Using linearity of $F$, we get, $(F(f))(g) = F\left(\int_{\mathcal{M}} f(y) u_y \omega^{\mathcal{M}}(y)\right)(g) = \int_{\mathcal{M}} f(y) F(u_y)(g) \omega^{\mathcal{M}}(y)$. Let us define,

$$\begin{aligned} w\left(g^{-1}.y\right) &:= \widetilde{w}_y(g) \\ &:= F(u_y)(g), \end{aligned} \tag{3}$$

for some $w \in S$ and $\widetilde{w}_y : G \to \mathbf{R}$. Clearly, $\widetilde{w}_y$ is linear since $F$ is. Below, we will show that $\widetilde{w}_y$ is equivariant and hence Eq. 3 is well-defined.

**Claim:** $\left(L_*^h \widetilde{w}_y\right)(g) = \widetilde{w}_y\left(h^{-1} g\right)$.

Since $F$ is equivariant, we get $\left(L_*^h F(u_y)\right)(g) = (F(u_y))(h^{-1} g)$. Now, from the definition in Eq. 3, we have $\left(L_*^h \widetilde{w}_y\right)(g) = \left(L_*^h F(u_y)\right)(g)$ and $\widetilde{w}_y(h^{-1} g) = (F(u_y))(h^{-1} g)$, hence, our claim holds.

Now, replacing $F(u_y)(g)$ with $w(g^{-1}.y)$ we get the desired result. ∎

The above theorem is analogous to the result in vector spaces on linear equivariant systems being fully characterized by the covolution, except in this case, the convolution is replaced by a correlation. A corollary of the above theorem is the following.

**Corollary 1.** *Let $F : S \to U$ be a function given by $f \mapsto (f \star w)$. Then, $F$ is equivariant.*

## 2.1 Orthogonal basis functions on homogeneous spaces

At the outset, we like to mention that the essential propositions/theorems and their proofs reported in this section are original to this work, although it is possible that they maybe found burried in the large body of harmonic analysis literature which we did not find despite immense efforts.

It is well-known that a homogeneous space $\mathcal{M}$ identified as $G/H$ together with the projection map $\Pi : G \to G/H$ is a principal bundle with $H$ as the fiber. In this section, we will show induced Fourier basis from the orthogonal basis on groups. In order to define the basis, we first note that the set of all square-integrable functions on $\mathcal{M}$, denoted by $L^2(\mathcal{M}, \mathbf{R})$ forms a vector space. We will equip this vector space with an inner product $\langle,\rangle : \mathcal{M} \times \mathcal{M} \to \mathbf{R}$ with respect to $\omega^\mathcal{M}$ as $\langle f_1, f_2 \rangle \mapsto \int_\mathcal{M} f_1(x) f_2(x) \omega^\mathcal{M}(x)$.

Now, let $\{v_\alpha : G \to \mathbf{R}\}$ be the set of orthogonal basis of $L^2(G, \mathbf{R})$. Using the principal bundle structure of the homogeneous space, we will get a set of basis on $L^2(\mathcal{M}, \mathbf{R})$ as follows:

**Proposition 2.** $\{\widetilde{v}_\alpha : \mathcal{M} \to \mathbf{R}\}$ *are linearly independent where,* $\widetilde{v} = v\widetilde{\pi}$, *where* $\widetilde{\pi}(x) = g$ *iff* $x = \pi(g)$.

**Proposition 3.** $\{\widetilde{v}_\alpha\}$ *as defined above span* $L^2(\mathcal{M}, \mathbf{R})$.

Using the above two propositions, we can show that $\{\widetilde{v}_\alpha\}$ is a set of basis of $L^2(\mathcal{M}, \mathbf{R})$ The following proposition shows that $\{\widetilde{v}_\alpha\}$ form a set of orthogonal basis.

**Proposition 4.** $\{\widetilde{v}_\alpha\}$ *as defined above is a set of orthogonal basis.*

This concludes that from the orthogonal basis on $G$, we get an induced orthogonal basis on $\mathcal{M}$ (please see the supplementary section for the proofs of all the propositions). We will use the Fourier basis on $G$ and the induced orthogonal basis on $\mathcal{M}$ to compute the correlation operation. In the next section, we will give an end-to-end trainable network architecture for diffusion MR data (identified as functions on homogeneous space). *To the best of our knowledge this is the first work to propose an end-to-end trainable network for dMRI data classification.*

## 3 An end-to-end trainable network architecture for dMRI data classification

dMRI is a non-invasive diagnostic imaging technique that allows one to infer the axonal connectivity within the imaged tissue by sensitizing the MR signal to water diffusion in imaged tissue [1]. In dMRI, most applications rely on the fundamental relationship between the MR signal measurement $S(\mathbf{q})$ and the average particle displacement density function (also known the ensemble average propagator or EAP) $P(\mathbf{r})$ at a voxel, $\mathbf{x} = (x, y, z)^t$. As already described before, the natural mathematical space for describing the data domain for $S(\mathbf{q})$ as well as the EAP, $P(\mathbf{r})$, is $\mathbf{S}^2 \times \mathbf{R}^+$.

### 3.1 Network architecture

In each voxel of the dMRI scan, the signal is acquired as a real number along each magnetic field direction over a hemi-sphere of directions in 3D. Hence, in each voxel, we have a function $f : \mathbf{S}^2 \times \mathbf{R}^+ \to \mathbf{R}$. As described earlier, we will use the well known SHORE basis [21, 12] to represent each function. Our proposed network architecture has two components, to extract *intra-voxel features* and *inter-voxel features* respectively. Below, we will describe both of these layers separately. A figure dpicting the HCNN architecture is included in the supplementary material.

### 3.1.1 Extracting intra-voxel features

In order to extract intra-voxel features, we will treat each voxel independently. As mentioned before, in each voxel we have a function $f : \mathbf{S}^2 \times \mathbf{R}^+ \to \mathbf{R}$. Since $\mathbf{S}^2 \times \mathbf{R}^+$ is a Riemannian homogeneous space (endowed with the product metric), we will use a sequence of correlation layers (with non-linearity within) to extract features which are *equivariant* to the action of $\mathsf{SO}(3) \times (\mathbf{R} \setminus \{0\})$. The architecture to extract this intra-voxel features consists of three layers described below. For simplicity of notations, we will use $\mathcal{N}$ to denote $\mathbf{S}^2 \times \mathbf{R}^+$ and $G$ to denote $\mathsf{SO}(3) \times (\mathbf{R} \setminus \{0\})$.

**Correlation on $\mathcal{N}$ - $\mathsf{Corr}^{\mathcal{N}}(f,w)$:** Let $f \in L^2(\mathcal{N}, \mathbf{R})$ be the input function and $w \in L^2(\mathcal{N}, \mathbf{R})$ be the mask. Then, $\mathsf{Corr}^{\mathcal{N}}(f,w) := (f \star w) : H \to \mathbf{R}$. We have shown in Corollary 1, that $\mathsf{Corr}^{\mathcal{M}}(f,w)$ is equivariant to action of $G$. Hence, we can use $\mathsf{Corr}^{H}(f,w)$ layer as the next layer.

**Correlation on $H$ - $\mathsf{Corr}^{H}(f,w)$:** Let $f \in L^2(H, \mathbf{R})$ be the input function and $w \in L^2(H, \mathbf{R})$ be the mask. Then, $\mathsf{Corr}^{H}(f,w) := (f \star w) : H \to \mathbf{R}$. Clearly, $\mathsf{Corr}^{H}(f,w)$ is equivariant to action of $H$. *Since this is an operation equivariant to $H$, we can cascade $\mathsf{Corr}^H(f,w)$.*

In order to use nonlinearity between two layers, we will define a Lie-Algebra based ReLU. Note that $H$ is a Lie group, hence there is a associated Lie algebra $\mathfrak{H}$ [14]. Since $\mathfrak{H}$ is a vector space, we define the standard ReLU on $\mathfrak{H}$. Note that, we will assume that $H$ is connected, hence, any point $h \in H$ can be parametrized by $(\mathfrak{h}_1, \cdots \mathfrak{h}_k) \in \mathfrak{H}^k$, for some finite $k$. Since each $\mathfrak{h}_i$ can be identified with a point on $\mathbf{R}^m$, where $m$ is the dimension of $H$, we apply ReLU on $\{\mathfrak{h}_i\}$ component-wise. After applying ReLU, we use the Riemannian exponential map [8] to map the result on to $H$. This layer is denoted by $H$-**ReLU**.

We will use a cascade of these layers to extract features – from each voxel independently – that are equivariant to the action of $H$. Observe that this equivariance property is natural in the context of dMRI data. Since in each voxel of the dMRI data, the signal is accquired in different directions (in 3D), we want the features to be equivariant to the 3D rotations and scaling (given by $H = \mathsf{SO}(3) \times (\mathbf{R} \setminus \{0\}))$. Thus, our formulation extracts features which are natural to the dMRI data.

### 3.1.2 Extracting inter-voxel features

After the extraction of the intra-voxel features (which are equivariant to the action of $H$), we want to derive features based on the interactions between the neighboring voxels. We will use a cascade of standard convolution and ReLU layers to capture the interaction between the equivariant intra-voxel features. This process yields features capturing the interactions between intra-voxel features over a spatial neighborhood. One can of course treat the intra-voxel features as a bag of features and use a fully connected layer. But, we will show using permutation testing in Section 4 that capturing interaction between features across voxels gives statistically significant results, while without this layer the result is not statistically significant. We will give the detail about this statistical testing in the experimental section. We will call this network architecture dMR-HCNN (abbreviation for homogeneous CNN for diffusion MR data).

## 4 Experiments

In this section, we present two sets of experiments for dMRI data, (i) a real data experiment that involves classification of dMRI brain scans acquired from a cohort of Parkinson Disease (PD) patients and control subjects, (ii) synthesized data to perform classification of data drawn from Gaussian densities on $P_3$.

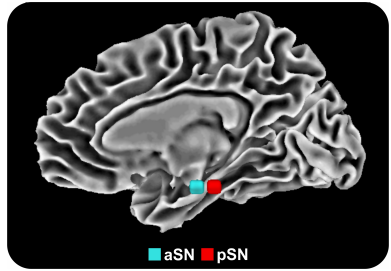

Figure 1: A sample $S(\mathbf{0})$ image with overlayed ROIs

For real dMRI data, we used two kinds of representations, the EAP, $P(\mathbf{r})$, and the raw signal, $S(\mathbf{q})$. The data pool consists of dMRI (human) brain scans acquired from 50 PD patients and 44 controls. All images were collected using a 3.0 T MR scanner (Philips Achieva) and 32-channel quadrature volume head coil. movement was minimized by foam padding within The parameters of the diffusion imaging acquisition sequence were as follows: gradient directions = 64, b-values = 0/1000 s/mm2, repetition time =7748 ms, echo time = 86 ms, flip angle = 90°, field of view = 224 × 224 mm, matrix size = 112 × 112, number of contiguous axial slices = 60, slice thickness = 2 mm, and SENSE factor P = 2.

From each subject, the left/right anterior and posterior substantia nigra (aSN and pSN respectively) were manually segmented by an expert. We restrict our attention to these 4 regions-of-interest (ROIs) for the classification task as they are known to be affected most by PD. Eddy current correction was applied to each data set by using standard motion correction techniques. In Fig. 1 depicts an example of $S_0$ (zero magnetic gradient) image in the Montreal Neurological Institute (MNI) standard coordinate space overlayed with two of the ROIs.

From each dMRI scan, we extracted EAP fields of dimension $112 \times 112 \times 60 \times 324$ (see [5] for details). Below, we provide the experimental details of classification for both these representations. We selected 85 subjects at random to train on and the remaining 9 were used for testing.

### 4.1 Classification on real dMRI data

For each ROI, we extracted the intra-voxel features using the dMR-HCNN with weights shared across voxels in the same ROI. The network architecture to extract intra-voxel features is as follows $\mathsf{Corr}^{\mathcal{N}} \to \text{H-ReLU} \to \mathsf{Corr}^H \to \text{H-ReLU} \mathsf{Corr}^H$ with the number of channels being 5, 10 and 15 respectively. Furthermore, we use batch-normalization after each convolution layer. Then, for each ROI, we extracted inter-voxel features using two $2 \times 2$ convolution layers with number of channels 20 and 25 respectively. We used $2 \times 2$ standard batch-normalization, max-pool and ReLU in between.

After, extracting inter voxel features, we combined features from 4 ROIs using a log-softmax fully connected layer with negative log-likelihood loss at the end. We used the SGD as the optimizer with initial step size of 0.1 with a step decay learning rate update. The total number of parameters for this network is 32482. We trained this model for 94 epochs and obtained 95.24% training and 88.88% testing accuracy. The total training time is 9328.8 seconds. For the given data, the above testing accuracy implies a mis-classification of one test data sample.

## 4.2 Classification using Ensemble Average Propagator (EAP)

As in the dMRI raw signal-based classification experiment, we extracted intra- and inter-voxel features for this EAP-based classification experiment. The parameters are shared within an ROI but are different across different ROIs. The network architecture to extract intra-voxel features is same as in the previous experiment with the number of channels set to 15, 30 and 45 respectively. The inter-voxel feature extractor has number of channels set to 50 and 55 respectively. Because of the large dimension of this EAP representation, this architecture has 217862 parameters, which is 6 times more than that in the previous experiment. We used the same optimizer and learning rate scheme as before and after 200 epochs we obtained 95.24% training and 88.88% testing accuracy. The total training time is 31271.9 seconds. From a classification accuracy view point, the performance achieved is similar to the raw-signal based experiment but at the expense of several fold increase in the number of parameters.

## 4.3 Permutation testing for statistical significance of inter-voxel features

In this section, we present a Hotelling $T^2$ statistic and use this test to assess the statistical significance of group differences. In order to achieve this, we perform a permutation test. Since it is difficult to formulate a parametric permutation test for this data, we use a non-parametric permutation test instead. The steps involved in performing the permutation test are as follows: (i) compute the $t^2$ statistic. (2) randomly permute the data between PD and control groups, and then compute $t_i^2$. (3) Repeat step (2) $50,000$ times and report the p-value as the fraction of times $t_i^2 > t^2$. The resultant p-value can be interpreted as the probability of finding a larger group difference by randomly permuting the data. We will reject the null hypothesis that there is no difference between the group means with 5% significance. The p-values for both dMR signal and EAP-based representations for inter- and intra-voxel features are reported in Table 1. We computed the p-values for each ROI independently.

| Mode | Intra/Inter | ROI | | | |
|------|-------------|-----|-----|-----|-----|
|      |             | aSN (L) | aSN (R) | pSN (L) | pSN (R) |
| dMR  | Intra | 0.39 | 0.52 | 0.45 | 0.97 |
| dMR  | Inter | **0.00** | **0.00** | **0.00** | **0.00** |
| EAP  | Intra | 0.44 | 0.82 | 0.39 | 0.65 |
| EAP  | Inter | **0.01** | 0.22 | 0.32 | 0.61 |

Table 1: p-values for permutation testing using dMR and EAP

In the above table, aSN/pSN (L/R) represent the left/right anterior/posterior substantia nigra respectively. By examining the p-values in Table 1, we can see that for both EAP and the raw dMR signal, intra-voxel features are not statistically significant in finding the group difference, while after examining the interaction between $H$-equivariant features, we can reject the null hypothesis. This justifies the necessity for inter-voxel feature layer. Furthermore, we used a fully connected layer after extracting intra-voxel features and obtain around 50% classification accuracy (i.e., uniform class probabilities). This in conjunction with the hypothesis testing described above indicates the usefulness of inter-voxel features. But, one may wonder about the usefulness of intra-voxel features, hence we used a 3-dimensional spatial convolution on the dMR signal to obtain around 66.67% accuracy. This clearly indicates the importance of both inter- and intra-voxel layers.

Furthermore, the results in Table 1, indicate that though using the dMR raw signal representation, features extracted from all the 4 ROIs are statistically significant, using EAP only yields a subset of the 4 ROIs, i.e., ROI aSN(L) is statistically significant. This testing implies that while using EAPs, features from left aSN suffice to find significant group difference. We suspect that this maybe because EAPs are more expressive than raw the raw dMR signal. Further investigation involving larger population studies are necessary to draw conclusive inferences in this context. This will be the focus of our future efforts.

### 4.4 Classification of data on $P_3$

In this section, we first describe the process of synthesizing functions $f : P_3 \to [0, 1]$. In this experiment, we generated data samples drawn from distinct Gaussian distributions defined on $P_3$ [4]. Let $\mathcal{X}$ be a $P_3$ valued random variable that follows $\mathcal{N}(M, \sigma)$, then, the p.d.f. of $X$ is given by $f_\mathcal{X}(X) = \frac{1}{C(\sigma)} \exp\left(-\frac{d^2(M,X)}{2\sigma^2}\right)$ [4]. We first chose two sufficiently spaced apart location parameters $M_1$ and $M_2$ and then for the $i^{th}$ class we generate Gaussian distributions with location parameters that are perturbations of $M_i$ and with variance 1. This gives us two clusters in the space of Gaussian densities on $P_3$, which we will classify using our proposed HCNN. In this case, the network architecture is given by: $\mathsf{Corr}^{P_3} \to \mathsf{GL}(3)\text{-ReLU} \to \mathsf{Corr}^{\mathsf{GL}(3)} \to \mathsf{GL}(3)\text{-ReLU} \to \mathsf{Corr}^{\mathsf{GL}(3)} \to \mathsf{GL}(3)\text{-ReLU} \to \mathsf{FC}$.

The data consists of 500 samples from each class, where each sample is drawn from a Gaussian distribution on $P_3$. Our implementation of the HCNN applied to this data yields a training accuracy of 92.5% (with stdev. 0.03) and a testing accuracy of 86.5% (with stdev. 0.02) in a ten-fold partition of the data. In most deep learning applications, one is used to seeing a high classification accuracy, but we believe that this can be achieved here as well by increasing the number of layers and possibly overfitting the data. The purpose of this synthetic experiment was not to seek an "optimal" classification accuracy but to provide a flexible framework which if "optimally" tuned can yield a good testing accuracy for data whose domain is a non-compact Riemannian homogeneous space.

# 5  Conclusions

In this paper, we presented a novel generalization of the CNN to cope with data whose domain is a Riemannian homogeneous space. We call our network a homogeneous CNN abbreviated HCNN. The salient contributions of our work are: (i) Linear group equivariance systems are shown to be characterized by correlation of functions on the domain manifold and vice-versa. This is fundamental to the characterization of all linear group equivariant systems and parallels the widely used result in linear system theory for vector spaces. (ii) As a corrolary, we prove the equivariance of the correlation operation to group actions admitted by Riemannian homogeneous manifolds. (iii) We present the first end-to-end deep network architecture for classification of diffusion magnetic resonance image (dMRI) scans acquired from a cohort of 44 Parkinson Disease patients and 50 control/normal subjects. (iv) A proof of concept experiment involving synthetic data generated on $P_3$ is presented to demonstrate the applicability of our network to other types of homogeneous spaces. Our future work will focus on testing on dMRI brains scans from a larger populations of PD patients and control subjects.

# References

[1] Peter J Basser, James Mattiello, and Denis LeBihan. Mr diffusion tensor spectroscopy and imaging. *Biophysical journal*, 66(1):259–267, 1994.

[2] Michael M. Bronstein, Joan Bruna, Yann LeCun, Arthur Szlam, and Pierre Vandergheynst. Geometric deep learning: Going beyond euclidean data. *IEEE Signal Process. Mag.*, 34(4):18–42, 2017.

[3] Paul T. Callaghan. *Principles of nuclear magnetic resonance microscopy*. Oxford University Press on Demand, 1993.

[4] Guang Cheng and Baba C Vemuri. A novel dynamic system in the space of spd matrices with applications to appearance tracking. *SIAM journal on imaging sciences*, 6(1):592–615, 2013.

[5] Guang Cheng, Baba C Vemuri, Paul R Carney, and Thomas H Mareci. Non-rigid registration of high angular resolution diffusion images represented by gaussian

mixture fields. In *International Conference on Medical Image Computing and Computer-Assisted Intervention*, pages 190–197. Springer, 2009.

[6] T.S. Cohen, M. Geiger, J. Koehler, and M. Welling. Convolutional networks for spherical signals. In *Proceedings of ICML*. JMLR, 2017.

[7] T.S. Cohen, M. Geiger, J. Koehler, and M. Welling. Spherical CNNs. In *Proceedings of ICLR*. JMLR, 2018.

[8] Manfredo Perdigao Do Carmo. *Riemannian geometry*. Birkhauser, 1992.

[9] Simoncelli E., Freeman W. T., Adleson E.H., and Heeger D. Shiftable multiscale transforms. *IEEE Transactions of Information Theory*, 38:587–607, 1992.

[10] Worrall Daniel E, Garbin Stephan J, Turmukhambetov Daniyar, , and Brostow Gabriel J. Harmonic networks: Deep translation and rotation equivariance. In *Proceedings of the IEEE CVPR*, pages 5026–5037. IEEE, 2017.

[11] Carlos Esteves, Christine Allen-Blanchette, Xiaowei Zhou, and Kostas Daniilidis. Polar transformer networks. *arXiv preprint arXiv:1709.01889*, 2017.

[12] Rutger HJ Fick, Demian Wassermann, Emmanuel Caruyer, and Rachid Deriche. MAPL: Tissue microstructure estimation using laplacian-regularized MAP-MRI and its application to hcp data. *NeuroImage*, 134:365–385, 2016.

[13] Freeman W. H. and Adelson E. H. The design and use of steerable filters for image analysis. *IEEE Transactions on PAMI*, 13:891–906, 1991.

[14] Brian Hall. *Lie groups, Lie algebras, and representations: an elementary introduction*, volume 222. Springer, 2015.

[15] Sigurdur Helgason. *Differential geometry and symmetric spaces*, volume 12. Academic press, 1962.

[16] Max Jaderberg, Karen Simonyan, Andrew Zisserman, et al. Spatial transformer networks. In *Advances in neural information processing systems (NIPS)*, pages 2017–2025, 2015.

[17] Risi Kondor and Shubhendu Trivedi. On the generalization of equivariance and convolution in neural networks to the action of compact groups. *arXiv preprint arXiv:1802.03690*, 2018.

[18] Alex Krizhevsky and Geoffrey Hinton. Learning multiple layers of features from tiny images. Technical report, Tech. Report, University of Toronto, 2009.

[19] Yann LeCun, Léon Bottou, Yoshua Bengio, and Patrick Haffner. Gradient-based learning applied to document recognition. *Proceedings of the IEEE*, 86(11):2278–2324, 1998.

[20] Jason D McEwen and Boris Leistedt. Fourier-laguerre transform, convolution and wavelets on the ball. *arXiv preprint arXiv:1307.1307*, 2013.

[21] Evren Özarslan, Cheng Guan Koay, Timothy M Shepherd, Michal E Komlosh, M Okan İrfanoğlu, Carlo Pierpaoli, and Peter J Basser. Mean apparent propagator (MAP) MRI: a novel diffusion imaging method for mapping tissue microstructure. *NeuroImage*, 78:16–32, 2013.

[22] Perona P. Steerable-scalable kernels for edge detection and junction analysis. In *Proceedings of ECCV*, pages 3–18. Springer, 1992.